\pdfoutput=1
\documentclass[11pt,a4paper]{article}
\usepackage[hyperref]{acl2018}
\usepackage{times}
\usepackage{latexsym}
\usepackage{graphicx}
\usepackage{url}
\usepackage{amssymb}
\usepackage{color}
\usepackage{soul}
\usepackage{floatrow}

\aclfinalcopy 

\title{Compositional Representation of Morphologically-Rich  Input\\ for Neural Machine Translation}

\author{Duygu Ataman \\
  FBK, Trento, Italy \\
  University of Trento, Italy \\
  {\tt ataman@fbk.eu} \\\And
  Marcello Federico \\
  MMT Srl, Trento, Italy \\
  FBK,  Trento, Italy \\
  {\tt federico@fbk.eu} \\}

\date{}

\begin{document}
\maketitle
\begin{abstract}
  Neural machine translation (NMT) models are typically trained with fixed-size input and output vocabularies, which creates an important bottleneck on their accuracy and generalization capability. As a solution, various studies proposed segmenting words into sub-word units and performing translation at the sub-lexical level. However, statistical word segmentation methods have recently shown to be prone to morphological errors, which can lead to inaccurate translations. In this paper, we propose to overcome this problem by replacing the source-language embedding layer of NMT with a bi-directional recurrent neural network that generates compositional representations of the input at any desired level of granularity. We test our approach in a low-resource setting with five languages from different morphological typologies, and under different composition assumptions. By training NMT to compose word representations from character n-grams, our approach consistently outperforms (from 1.71 to 2.48 BLEU points) NMT learning embeddings of statistically generated sub-word units.
\end{abstract}

\section{Introduction}

An important problem in neural machine translation (NMT) is translating infrequent or unseen words. The reasons are twofold: the necessity of observing many examples of a word until its input representation (embedding) becomes reliable, and the computational requirement of limiting the input and output vocabularies to few tens of thousands of words. These requirements eventually lead to coverage issues when dealing with low-resource and/or morphologically-rich languages, due to their high lexical sparseness. To cope with this well-known problem, several approaches have been proposed redefining the model vocabulary in terms of interior orthographic units compounding the words, ranging from character n-grams~\cite{ling2015character,costa2016character,lee2016fully,luong2016achieving} to statistically-learned sub-word units~\cite{sennrich2015neural,wu2016google,ataman}. 
While the former provide an ideal open vocabulary solution, they mostly failed to achieve competitive results. This might be related to the semantic ambiguity caused by solely relying on input representations based on character n-grams which are generally learned by disregarding any morphological information. In fact, the second approach is now prominent and has established a pre-processing step for constructing a vocabulary of sub-word units before training the NMT model. However, several studies have shown that segmenting words into sub-word units without preserving morpheme boundaries can lead to loss of semantic and syntactic information and, thus, inaccurate translations~\cite{niehues,ataman,pinnis,huck2017target,fraser2}.

In this paper, we propose to improve the quality of input (source language) representations of rare words in NMT by augmenting its {\em embedding layer} with a {\em bi-directional recurrent neural network} (bi-RNN), which can learn compositional input representations at different levels of granularity. Compositional word embeddings have recently been applied in language modeling and obtained successful results \cite{vania2017characters}. The apparent advantage of our approach is that by feeding NMT with simple character n-grams, our bi-RNN can potentially learn the morphology necessary to create word-level representations of the input language directly at training time, thus, avoiding the burden of a separate and sub-optimal word segmentation step. We compare our approach against conventional embedding-based representations learned from statistical word segmentation in a public evaluation benchmark, which provides low-resource training conditions by pairing English with five morphologically-rich languages: Arabic, Czech, German, Italian and Turkish, where each language represents a distinct morphological typology and language family. The experimental results show that our compositional input representations lead to significantly and consistently better translation quality in all language directions.

\section{Neural Machine Translation}
\label{nmt}

In this paper, we use the NMT model of Bahdanau et al.~\shortcite{bahdanau2014neural}. The model essentially estimates the conditional probability of translating a source sequence $x = (x_1, x_2, \ldots x_m)$ into a target sequence $y = (y_1, y_2, \ldots y_l)$, using the decomposition
\begin{equation}
\begin{small}
\label{nmt1}
   p(y|x) = \\
   \prod_{i=1}^l p(y_j|y_{i-1},..,y_0,x_{m-1},..,x_1)
\end{small}
\end{equation}

\noindent
The model is trained by maximizing the log-likelihood of a parallel training set via stochastic gradient descent \citep{sgd} and the backpropagation through time  \citep{werbos1990backpropagation} algorithms.

The inputs of the network are {\it one-hot} vectors, which are binary vectors with a single bit set to 1 to identify a specific word in the vocabulary. Each one-hot vector is then mapped to an {\it embedding}, a distributed representation of the word in a lower dimension but a more dense continuous space. From this input, a representation of the whole input sequence is learned using a bi-RNN, the \textit{encoder}, which maps \textit{x} into \textit{m} dense sentence vectors corresponding to its hidden states. Next, another RNN, the \textit{decoder}, predicts each target token $y_{i}$ by sampling from a distribution computed from the previous target token $y_{i-1}$, the previous decoder hidden state, and the \textit{context vector}. The latter is a linear combination of the encoder hidden states, whose weights are dynamically computed by a feed-forward neural network called \textit{attention model} \cite{bahdanau2014neural}. The probability of generating each target word $y_{j}$ is normalized via a softmax function.

Both the source and target vocabulary sizes play an important role in terms of defining the complexity of the model. In a standard architecture, like ours, the source and target embedding matrices actually account for the vast majority of the network parameters. The vocabulary size also plays an important role when translating from and to low-resource and morphologically-rich languages, due to the sparseness of the lexical distribution. Therefore, a conventional approach has now become to compose both the source and target vocabularies of sub-word units generated through statistical segmentation methods \cite{sennrich2015neural,wu2016google,ataman}, and performing NMT by directly learning embeddings of sub-word units. A popular one of these is the Byte-Pair Encoding (BPE) method~\cite{gage1994new, sennrich2015neural}, which finds the optimal description of a corpus vocabulary by iteratively merging the most frequent character sequences. A more recent approach is the Linguistically-Motivated Vocabulary Reduction (LMVR) method~\cite{ataman}, which similarly generates a new vocabulary by segmenting words into sub-lexical units based on their likeliness of being morphemes and their morphological categories. A drawback of these methods is that, as pre-processing steps to NMT, they are not optimized for the translation task. Moreover, they can suffer from morphological errors at different levels, which can lead to loss of semantic or syntactic information. 





\begin{figure*}[t]
\floatbox[{\capbeside\thisfloatsetup{capbesideposition={right,top},capbesidewidth=4cm}}]{figure}[\FBwidth]
{\caption{Translation of the Italian sentence \textit{tornai a casa} (\textit{I came home}) with a word-level representation composed from character trigrams.}\label{fig:fig1}}
{\includegraphics[width=11.5cm]{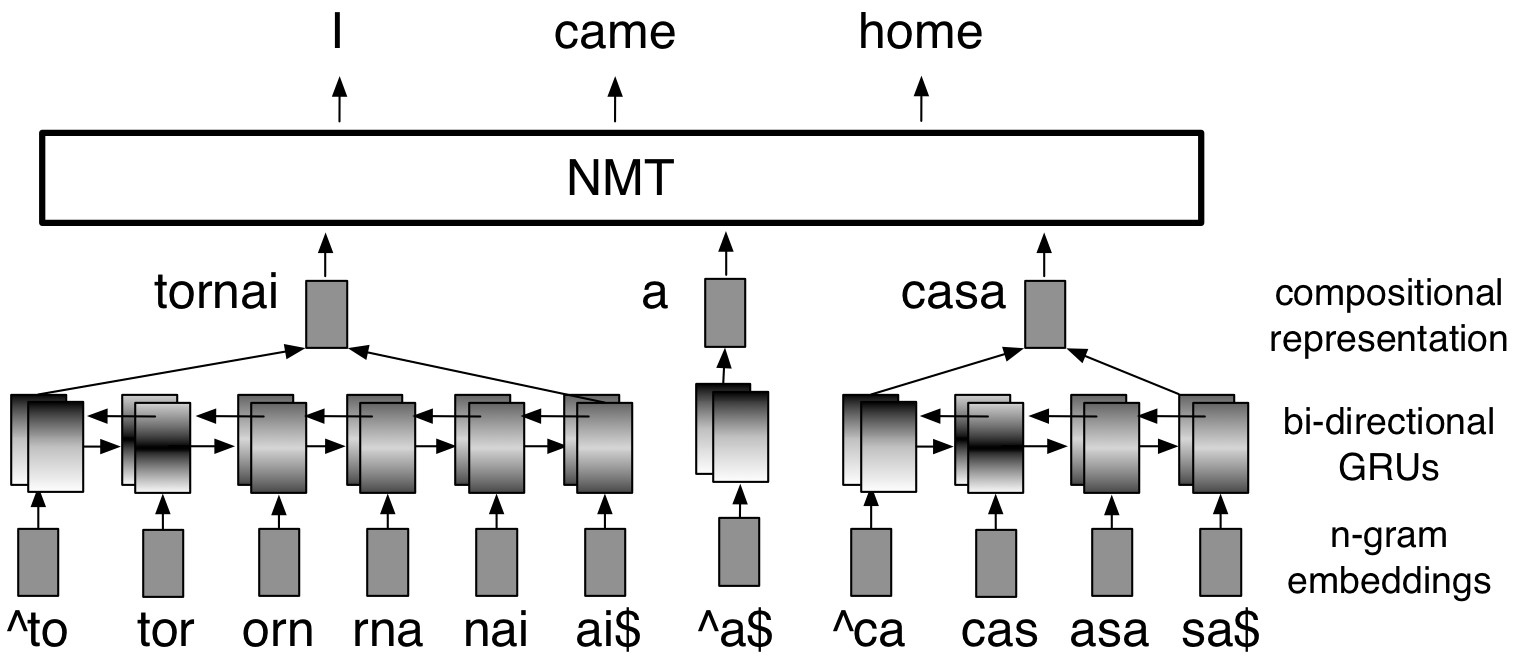}}
\end{figure*}

\section{Learning Compositional Input Representations via bi-RNNs}
\label{method}

In this paper, we propose to perform NMT from input representations learned by composing smaller symbols, such as character n-grams~\cite{ling2015finding}, that can easily fit in the model vocabulary. This composition is essentially a function which can establish a mapping between combinations of ortographic units and lexical meaning, that is learned using the bilingual context so that it can produce representations that are optimized for  machine translation. 

In our model (Figure \ref{fig:fig1}), the one-hot vectors, after being fed into the embedding layer, are processed by an additional \textit{composition layer}, which computes the final input representations passed to the encoder to generate translations. For learning the composition function, we employ a bi-RNN. Hence, by encoding each interior unit inside the word, we hope to capture important cues about their functional role, \textit{i.e.} semantic or syntactic contribution to the word. We implement the network using gated recurrent units (GRUs) \cite{cho2014properties}, which have shown comparable performance to long-short-term-memory units \cite{hochreiter1997long}, whereas they provide much faster computation. As a minimal set of input symbols required to cope with contextual ambiguities, we opt to use intersecting sequences of character trigrams, as recently suggested by Vania and Lopez \shortcite{vania2017characters}. 

Given a bi-RNN with a forward ($f$) and backward ($b$) layer, the input representation $\mathbf{w}$ of a token of $t$ characters is computed from the hidden states $\mathbf{h}_t^f$ and $\mathbf{h}_b^0$, \textit{i.e.} the final outputs of the forward and backward RNNs, as follows:

\begin{equation}
    \mathbf{w} = \mathbf{W}_f  \mathbf{h}_f^t + \mathbf{W}_b · \mathbf{h}_b^0 + \mathbf{b}
\end{equation}

where $\mathbf{W}_f$ and $\mathbf{W}_b$ are weight matrices associated to each RNN and $\mathbf{b}$ is a bias vector~\cite{ling2015finding}. These parameters are jointly learned together with the internal parameters of the GRUs and the input token embedding matrix while training the NMT model. For an input of $m$ tokens, our implementation increases the computational complexity of the network by $O(K t_{\max} m)$, where $K$ is the bi-RNN cost and $t_{\max}$ is the maximum number of symbols per word. However, since computation of each input representation is independent, a parallelised implementation could cut the overhead down to $O(K t_{\max})$.


\section{Experiments}

We test our approach along with statistical word segmentation based open vocabulary NMT methods in an evaluation benchmark simulating a low-resource translation setting pairing English ({\it En}) with five languages from different language families and morphological typologies: Arabic ({\it Ar}), Czech ({\it Cs}), German ({\it De}), Italian ({\it It}) and Turkish ({\it TR}). The characteristics of each language are given in Table \ref{language-table}, whereas Table \ref{tab:data} presents the statistical properties of the training data. We train our NMT models using the TED Talks corpora \citep{tedtalks} and test them on the official data sets of IWSLT\footnote{The International Workshop on Spoken Language Translation with shared tasks organized between 2003-2017.} \cite{mauro2017overview}. 

\begin{table}[h]
\begin{centering}
    \begin{tabular}{c|c|c}
    \hline \hline
    \textbf{Language} & \textbf{Morphological} & \textbf{Morphological}  \\
    &  \textbf{Typology} & \textbf{Complexity}\\
    \hline \hline
    Turkish & {\it Agglutinative} & {\it High} \\
    \hline
     Arabic & {\it Templatic} & {\it High} \\
     \hline
     Czech & {\it Fusional,} &  {\it High} \\
     & {\it Agglutinative} & \\
     \hline
     German & {\it Fusional} & {\it Medium} \\
     \hline
     Italian & {\it Fusional} & {\it Low} \\
     \hline \hline
    \end{tabular}
  \caption{The languages evaluated in our study and their morphological characteristics.}
  \label{language-table}
  \end{centering}
\end{table}

\begin{table}[h]
  \begin{centering}
    \begin{tabular}{c|cc|cc}
    \hline \hline
    \textbf{Language} & \multicolumn{2}{c|}{\textbf{\# tokens}} & \multicolumn{2}{c}{\textbf{\# types}}\\
    {\bf Pair} & {\bf Src} & {\bf Tgt} & {\bf Src} & {\bf Tgt} \\
    \hline \hline
    Tr - En & 2,7M & 2,0M & 171K & 53K \\
    \hline
    Ar - En & 3,9M & 4,9M & 220K & 120K \\
    \hline
    Cs - En & 2,0M & 2,3M & 118K & 50K \\
    \hline
    De - En & 4,0M & 4,3M & 144K & 69K \\
    \hline
    It - En & 3,5M & 3,8M & 95K & 63K \\
    \hline \hline
    \end{tabular}
    \caption{Sizes of the training sets and vocabularies in the TED Talks benchmark. Development and test sets are on average 50K to 100K tokens. (\textit{M}: Million, \textit{K}: Thousand.)}
    \label{tab:data}
  \end{centering}
\end{table}

The \textit{simple} NMT model constitutes the baseline in our study and performs translation directly at the level of sub-word units, which can be of four different types: characters, character trigrams, BPE sub-word units, and LMVR sub-word units. The \textit{compositional} model, on the other hand, performs NMT with input representations composed from sub-lexical vocabulary units. In our study, we evaluate representations composed from character trigrams, BPE, and LMVR units. In order to choose the segmentation method to apply on the English side (the output of NMT decoder), we compare BPE and LMVR sub-word units by carrying out an evaluation on the official data sets of Morpho Challenge 2010\footnote{Shared Task on Unsupervised Morphological Analysis, http://morpho.aalto.fi/events/morphochallenge.}\cite{kurimo2010morpho}. The results of this evaluation, as given in Table \ref{tab:morphochallenge}, suggest that LMVR seems to provide a segmentation that is more consistent with morpheme boundaries, which motivates us to use sub-word tokens generated by LMVR for the target side. This choice aids us in evaluating the morphological knowledge contained in input representations in terms of the translation accuracy in NMT. 

\begin{table}[b]
  \begin{centering}
    \begin{tabular}{c|c|c|c}
    \hline \hline
    \textbf{Method} & \textbf{Precision} & \textbf{Recall} & \textbf{F}$_1$ \textbf{Score}\\
    \hline \hline
    BPE  &  52.87 & 24.44 & 33.43 \\
    LMVR &  \textbf{70.22} & \textbf{55.66} & \textbf{62.10} \\
    \hline \hline
    \end{tabular}
    \caption{The performance of different segmentation models trained on the English portion of our benchmark in the Morpho Challenge shared task.}
    \label{tab:morphochallenge}
  \end{centering}
\end{table}

\begin{table*}[t]
\begin{center}
\begin{tabular}{c|c|c|c|c|c|c|c}
\hline \hline
{\bf Model} & {\bf Vocabulary } & \textbf{Input} & \multicolumn{5}{c}{\textbf{BLEU}} \\
& \textbf{Units} & \textbf{Representations} & {\bf Tr-En} & {\bf Ar-En} & {\bf Cs-En} & {\bf De-En} &  {\bf It-En} \\
\hline \hline
\textit{Simple} & Characters & Characters & 12.29 & 8.95 & 13.42 & 21.32 & 22.88 \\
& Char Trigrams & Char Trigrams & 16.13 & 11.91 & 20.87 & 25.01 & 26.68 \\
& Subwords (BPE) & Subwords (BPE) & 16.79 & 11.14 & 21.99 & 26.61 & 27.02 \\
& Subwords (LMVR) & Subwords (LMVR) & 17.82 & 12.23 & 22.84 & 27.18 & 27.34 \\
\hline
\textit{Composi-} & Char Trigrams & Subwords (BPE) & 15.40 & 11.50 & 21.67 & 27.05 & 27.80 \\
\textit{tional} & Char Trigrams & Subwords (LMVR) & 16.63 & 13.29 & 23.07 & 26.86 & 26.84 \\
& Char Trigrams & Words  & \textbf{19.53} & \textbf{14.22} & \textbf{25.16} & \textbf{29.09} & \textbf{29.82} \\
 & Subwords (BPE) & Words  & 12.64& 11.51 & 23.13 & 27.10 & 27.96\\
 & Subwords (LMVR) & Words & 18.90& 13.55& 24.31 & 28.07 & 28.83 \\
\hline \hline
\end{tabular}
\caption{Experiment results. Best scores for each translation direction are in bold font. All improvements over the baseline (simple model with BPE) are statistically significant (\textit{p-value}~$<$~$0.05$).}
\label{tab:results}
\end{center}
\end{table*}

The compositional bi-RNN layer is implemented in Theano \cite{theano} and integrated into the Nematus NMT toolkit \citep{nematus}. In our experiments, we use a compositional bi-RNN with 256 hidden units, an NMT model with a one-layer bi-directional GRU encoder and one-layer GRU decoder of 512 hidden units, and an embedding dimension of 256 for both models. We use a highly restricted dictionary size of 30,000 for both source and target languages, and train the segmentation models (BPE and LMVR) to generate sub-word vocabularies of the same size. We train the NMT models using the Adagrad \citep{duchi2011adaptive} optimizer with a mini-batch size of 50, a learning rate of 0.01, and a dropout rate of 0.1 (in all layers and embeddings). In order to prevent over-fitting, we stop training if the perplexity on the validation does not decrease for 5 epochs, and use the best model to translate the test set. The model outputs are evaluated using the (case-sensitive) BLEU \citep{bleu} metric and the Multeval ~\citep{multeval} significance test.

\section{Results}

\begin{table*}[h]
\begin{center}
\begin{small}

\begin{tabular}{c|c}
\hline \hline
\textbf{Input} & e comunque, em@@ ig@@ \textit{riamo} , circol@@ \textit{iamo} e mescol@@ \textit{iamo} cos\`i tanto che \\
\textbf{(Simple Model)} & non esiste pi\`u l' isolamento necessario affinch\'e avvenga un' evoluzione .\\
\hline
\textbf{NMT Output} & and anyway , \textit{we} repair, and \textit{we} mix so much that \\
\textbf{(Simple Model)} & there 's no longer the isolation that \textit{we} need to happen to make an evolution .\\
\hline
\textbf{Input } & e comunque, emigriamo, circoliamo e mescoliamo cos\`i tanto che \\
\textbf{(Compositional Model)}& non esiste pi\`u l' isolamento necessario affinch\'e avvenga un' evoluzione.\\
\hline
\textbf{NMT Output} & and anyway , \textit{we} migrate , circle and mix so much that \\
\textbf{(Compositional Model)}& there 's no longer the isolation necessary to become evolutionary .\\
\hline
\textbf{Reference} & and by the way , \textit{we} immigrate and circulate and intermix so much that  \\
& you can 't any longer have the isolation that is necessary for evolution to take place . \\
\hline \hline
\multicolumn{2}{c}{  } \\
\end{tabular}

\begin{tabular}{c|c}

\hline \hline
\textbf{Input} & ama asl{\i}nda bu resim tamamen , farkl{\i} \textit{yerlerin foto\u{g}raf@@ lar{\i}n{\i}n} \\
\textbf{(Simple Model)} & \textbf{birle\c{s}tir@@ il@@ mesiyle meydana geldi} . \\
\hline
\textbf{NMT Output} & but in fact , this picture \textbf{came up with} a completely \\
\textbf{(Simple Model)} & different \textit{place of photographs} . \\
\hline
\textbf{Input } & ama asl{\i}nda bu resim tamamen , farkl{\i} \textit{yerlerin foto\u{g}raflar{\i}n{\i}n}  \\
\textbf{(Compositional Model)}& \textbf{birle\c{s}tirilmesiyle meydana geldi} . \\
\hline
\textbf{NMT Output} & but in fact , this picture \textbf{came from collecting} \textit{pictures of} \\
\textbf{(Compositional Model)} & different \textit{places} . \\
\hline
\textbf{Reference} & but this image \textbf{is} actually entirely \textbf{composed of} \textit{photographs from} different \textit{locations} . \\
\hline \hline
\end{tabular}

\caption{Example translations with different approaches in \textit{Italian} (above) and \textit{Turkish} (below).}
\label{tab:examples}
\end{small}
\end{center}
\end{table*}

The performance of NMT models in translating each language using different vocabulary units and encoder input representations can be seen in Table \ref{tab:results}. 
With the simple model, LMVR based units achieve the best accuracy in translating all languages, with improvements over BPE by \textbf{0.85} to \textbf{1.09} BLEU points in languages with high morphological complexity (Arabic, Czech and Turkish) and \textbf{0.32} to \textbf{0.53} BLEU points in languages with low to medium complexity (Italian and German). This confirms our previous results in \cite{ataman2}. Moreover, simple models using character trigrams as vocabulary units reach much higher translation accuracy compared to models using characters, indicating their superior performance in handling contextual ambiguity. In the Italian to English translation direction, the performance of simple models using character trigrams and BPE sub-word units as input representations are almost comparable, showing that character trigrams can even be sufficient as the stand-alone vocabulary units in languages with low lexical sparseness. These findings suggest that each type of sub-word unit used in the simple model is specifically convenient for a given morphological typology. 

Using our compositional model improves the quality of input representations for each type of vocabulary unit, nevertheless, the best performance is obtained by using character trigrams as input symbols and words as input representations. The higher quality of these input representations compared to those obtained from sub-word units generated with LMVR suggest that our compositional model can learn morphology better than LMVR, which was found to provide comparable performance to morphological analyzers in Turkish to English NMT \cite{ataman}. Moreover, sample outputs from both models show that the compositional model is also able to better capture syntactic information of input sentences. Figure \ref{tab:examples} illustrates two example translations from Italian and Turkish. In Italian, the simple model fails to understand the common subject of different verbs in the sentence due to the repetition of the same inflective suffix after segmentation. In Turkish, the genitive case "yerlerin foto\u{g}raflar{\i}n{\i}n" (\textit{the photographs of places}) and the complex predicate "birle\c{s}tirilmesiyle meydana geldi" (\textit{is composed of}) are both incorrectly translated by the simple model. On the other hand, the compositional model is able to capture the correct sentence semantics and syntax in either case. These findings suggest that maintaining translation at the lexical level apparently aids the attention mechanism and provides more semantically and syntactically consistent translations. 
The overall improvements obtained with this model over the best performing simple model are \textbf{1.99} BLEU points in Arabic, \textbf{2.32} BLEU points in Czech, \textbf{1.91} BLEU points in German, \textbf{2.48} BLEU points in Italian and \textbf{1.71} BLEU points in Turkish to English translation directions. As evident from the significant and consistent improvements across all languages, our approach provides a more promising and generic solution to the data sparseness problem in NMT. 

\section{Conclusion}
In this paper, we have addressed the problem of translating infrequent and unseen words in NMT and proposed to solve it by replacing the conventional (source language) sub-word embeddings with input representations compositionally learned from character n-grams using a bi-RNN. Our approach showed significant and consistent improvements over a variety of languages with different morphological typologies, making it a competitive solution for NMT of low-resource and morphologically-rich languages. In the future, we plan to develop a more efficient implementation of our approach and to test its scalability on larger data sets. Our implementation and evaluation benchmark are available for public use.

\section*{Acknowledgments}
The authors would like to thank NVIDIA for their computational support that allowed to conduct this research.

\bibliography{acl2018}
\bibliographystyle{acl_natbib}

\appendix

\end{document}